\documentclass[letterpaper, 10 pt, conference]{ieeeconf}  % Comment this line out if you need a4paper

\IEEEoverridecommandlockouts                              % This command is only needed if 
\usepackage{graphicx}
\usepackage{booktabs}
\usepackage{amssymb}
\usepackage{amsmath}
\usepackage{algorithm}
\usepackage{algorithmicx}
\usepackage{algpseudocode}
\newcommand{\etal}{\textit{et al.}}
\newcommand{\ie}{\textit{i.e.}}
\newcommand{\eg}{\textit{e.g.}}

\usepackage{etoolbox}
\makeatletter
\patchcmd{\@makecaption}
  {\scshape}
  {}
  {}
  {}
\makeatother
\title{\LARGE \bf
PNS: Population-Guided Novelty Search for Reinforcement Learning in Hard Exploration Environments
}

\author{Qihao Liu$^{1,}$$^{2}$, Yujia Wang$^{2}$, and Xiaofeng Liu$^{2}$$^{*}$% <-this % stops a space
\thanks{This work was supported by the Natural Science Foundation of China (Grant: 11772187, 11802174). Work mainly performed at SJTU.}
\thanks{$^{1}$Laboratory for Computational Sensing and
Robotics, Johns Hopkins University, Baltimore, MD, USA.}%
\thanks{$^{2}$Shanghai Jiao Tong University, Shanghai, China. }
\thanks{$^{*}$Corresponding author: {\tt\small peterliuxiaofeng@sjtu.edu.cn}}%
}

\begin{document}

\maketitle
\thispagestyle{empty}
\pagestyle{empty}

%%%%%%%%%%%%%%%%%%%%%%%%%%%%%%%%%%%%%%%%%%%%%%%%%%%%%%%%%%%%%%%%%%%%%%%%%%%%%%%%
\begin{abstract}

Reinforcement Learning (RL) has made remarkable achievements, but it still suffers from inadequate exploration strategies, sparse reward signals, and deceptive reward functions. To alleviate these problems, a Population-guided Novelty Search (PNS) parallel learning method is proposed in this paper. In PNS, the population is divided into multiple sub-populations, each of which has one chief agent and several exploring agents. The chief agent evaluates the policies learned by exploring agents and shares the optimal policy with all sub-populations. The exploring agents learn their policies in collaboration with the guidance of the optimal policy and, simultaneously, upload their policies to the chief agent. To balance exploration and exploitation, the Novelty Search (NS) is employed in every chief agent to encourage policies with high novelty while maximizing per-episode performance. We apply PNS to the twin delayed deep deterministic (TD3) policy gradient algorithm. The effectiveness of PNS to promote exploration and improve performance in continuous control domains is demonstrated in the experimental section. Notably, PNS-TD3 achieves rewards that far exceed the SOTA methods in environments with sparse or delayed reward signals. We also demonstrate that PNS enables robotic agents to learn control policies directly from pixels for sparse-reward manipulation in both simulated and real-world settings.

\end{abstract}

%%%%%%%%%%%%%%%%%%%%%%%%%%%%%%%%%%%%%%%%%%%%%%%%%%%%%%%%%%%%%%%%%%%%%%%%%%%%%%%%
\section{INTRODUCTION}

RL is a type of machine learning algorithm that relies on an agent interacting with the environment and learning an optimal policy by trial and error. It has performed amazing feats in the fields of games, navigation, and control of physical systems. However, it still suffers from many tough problems. In many complex control tasks, especially in some challenging environments with spare reward signals or deceptive reward functions, greedy strategies will lead to limited exploration, and the RL algorithms based on these strategies often fail to learn a good policy~\cite{osband2016deep}. To deal with that, many previous works focus on using manually designed dense reward functions~\cite{zhu2019dexterous, gu2017deep}, or multi-goal-based training~\cite{andrychowicz2017hindsight,zhao2019curiosity}. But these methods may lead to suboptimal policies~\cite{riedmiller2018learning}, need domain knowledge~\cite{inoue2017deep}, and still require millions of training samples~\cite{andrychowicz2017hindsight}. We argue that one of the key ideas to solve this problem is to maintain efficient and meaningful exploration in these challenging environments.

\begin{figure}
  \centering
  \includegraphics[scale=1.2]{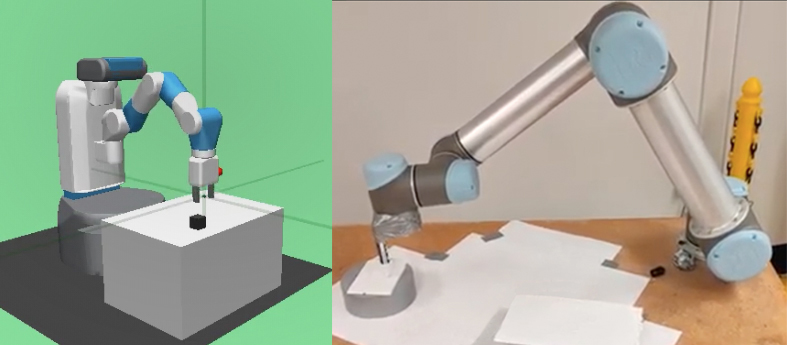}
  \caption{\textbf{P}opulation-Guided \textbf{N}ovelty \textbf{S}earch enables robotic agents to learn skills directly from pixels in sparse reward or delayed reward settings.}
  \setlength{\belowcaptionskip}{-1cm}
  \label{teaser}
\end{figure}

Up to now, a number of methods have been proposed to encourage exploration. A common idea is to encourage a single agent to explore the states that have rarely been explored. For example, Bellemare \etal~\cite{bellemare2016unifying} and Haber \etal~\cite{haber2018learning} quantify the difference between the current state and previous ones and provide intrinsic rewards to encourage exploration. Houthooft \etal~\cite{houthooft2016vime} and Pathak \etal~\cite{pathak2017curiosity} learn an environment dynamics model and encourage the agents to take actions that result in large updates to the dynamics model distribution. Those methods all approximate state visitation separately, may ultimately reduce to undirected exploration, and take the cost of high sample complexity. Another idea is to use multiple agents to explore the environment. For example, Jaderberg \etal~\cite{jaderberg2017population,jaderberg2019human} and Jung \etal~\cite{jung2020population} use population-based training (PBT) to perform exploration with a population of agents. Using multiple agents to explore environments is an effective method and works well in practice~\cite{gu2017deep,jaderberg2019human}, but existing methods do not make sufficient use of population-based exploration, and cannot learn efficiently in sparse reward settings.

In this paper, based on the PBT method, a Population-guided Novelty Search (PNS) parallel learning scheme for deep RL algorithms is proposed to encourage direct exploration and enable agents to learn skills in hard exploration environments. The scheme consists of multiple sub-populations, each sub-population includes several exploring agents that explore the environment and learn the policy, and one chief agent that evaluates the performance and the novelty of the policies learned by exploring agents and shares the current optimal policy with all sub-populations. At the beginning of training, exploring agents will be initialized using different values. Then, they will explore the environment and update their parameters independently with the guidance of the current optimal policy. Simultaneously, exploring agents will compare the performance of the current execution policy with the performance of the current optimal policy. If the former performs better than the latter, the policy parameters corresponding to the former are uploaded to the corresponding chief agents. At the same time, chief agents will evaluate the cumulative return and the novelty of the uploaded policy to decide whether to update the current optimal policy or not. The novelty evaluation draws on the approach of novelty search (NS). After the optimal policy is decided, the chief agents will share it with all sub-populations so that exploring agents can learn from it, which helps agents avoid local optima and accelerate training process. Meanwhile, the use of NS in chief agents provides extra information of the environment and encourages the exploring agents to explore the areas of parameter space that are not or rarely explored, thus promotes directed exploration and enables the agents to learn from sparse signals.

This paper makes the following contributions: (1) experimentally confirming that the combination of sub-population and NS in Deep RL can help the agents avoid local optima, promote directed exploration, and learn from sparse reward signals; (2) proposing PNS, which can be easily applied to any deep RL algorithm (\eg, TD3, PPO) and enhance learning performance; (3) proving that the proposed scheme performs better than SOTA parallel learning schemes, especially in environments with sparse or delayed reward signals. %We apply PNS to TD3, A2C, and PPO as our baseline algorithms. Our experiments on Atari, MuJoCo, and delayed MuJoCo show that the resultant new algorithms, PNS-TD3, PNS-A2C, and PNS-PPO, achieve higher performance, especially in hard exploration environments.

\section{Background and related work}
\label{Background}{\bf Distributed RL} Distributed learning is a commonly used speed-up method. It is based on the idea of parallel computing, and it can speed up the training with a small change to the existing reinforcement learning algorithm. Most of the existing distributed RL algorithms use multiple agents interacting with multiple copies of the same environment in parallel to generate more data sets, collect gradients of all agents through a central parameter server, and then use these gradients to update the parameter, including the asynchronous update of parameters~\cite{nair2015massively,mnih2016asynchronous,babaeizadeh2016reinforcement,stooke2018accelerated}, prioritized experience replay buffer~\cite{horgan2018distributed}, and V-trace to correct the distributional shift~\cite{espeholt2018impala,luo2019impact}. Different from our method, these distributed approaches use multiple actors to traverse different states in an asynchronous way to increase the sampling speed. The policy used by the actors responsible for sampling is a replica of the network, which does not take advantage of interactions among the populations and leads to the lack of strategy exploration capabilities. 

{\bf Population-Based Training (PBT)} PBT is another kind of parallel algorithm used for finding optimal parameters of a network model~\cite{jaderberg2017population}. The key idea is that multiple learners have different parameters and hyper-parameters. During training, PBT evaluates each learner’s performance first, then it selects and distributes the best parameters and hyper-parameters to other learners periodically. It is used in Quake III Arena~\cite{jaderberg2019human} and in AlphaStar's league~\cite{vinyals2019alphastar,arulkumaran2019alphastar}, which proves its ability and application prospect in large-scale training for solving hard exploration problems. But they only use PBT as a simple training method and during training, the parameter of the best learner is just copied to other learners, which sacrifices the diversity and exploration.  Recently, PBT is applied to Population-guided Parallel Policy Search (P3S) algorithm~\cite{jung2020population}. P3S uses the optimal policy to guide other learners' policy though an augmented loss function, but they do not consider the novelty of the policy and cannot learn efficiently in sparse reward settings.

{\bf Novelty Search (NS) } NS is a kind of algorithm to improve exploration. NS encourages population diversity by computing the novelty of the current policy with respect to previously executed policies and encourages the population distribution to move towards areas of parameter space with higher novelty. Conti \etal~\cite{conti2018improving} investigate how to combine NS with evolution strategies (ES). They add NS and quality diversity (QD) algorithms to ES to balance exploration and exploitation and improve the performance of ES on sparse and/or deceptive control tasks. However, NS occupies a lot of computer resources. While in the training process of RL algorithms, the parameters of the policy are updated at a high speed, so the novelty evaluation of each policy will greatly slow down the training speed. In this paper, we study how to incorporate the idea of NS into RL algorithms while 
reducing sacrificing data efficiency and saving the wall-clock time. We introduce NS and sub-population when learning the optimal policy, thus encouraging the population distribution to move towards areas that are not explored or rarely explored while maintaining a high score and training speed.

\section{Methods}
\label{methods}
\begin{figure}
  \centering
  \includegraphics[scale=0.8]{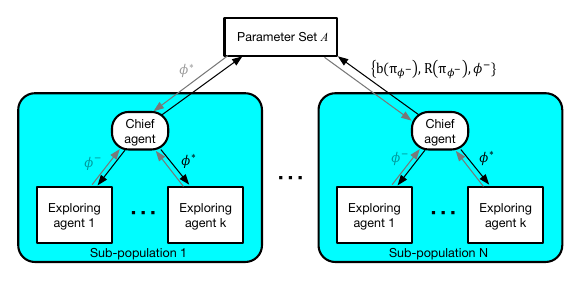}
  \caption{PNS consists of $n$ sub-populations, each of which includes $k$ exploring agents and one chief agent. Each exploring agent explores the environment independently with the guidance of its chief agent.}
  \label{figure_1}
\end{figure}
In this paper, a Population-guided Novelty Search (PNS) parallel learning scheme is proposed. As shown in Fig.\ref{figure_1}, our scheme consists of $n$ sub-populations, each of which includes $k$ exploring agents and one chief agent. All of the exploring agents learn the optimal policy using the same RL algorithm, such as TD3. Each of the exploring agents has an independent environment $\varepsilon^e$, which is a copy of the common environment $\varepsilon$, independent value function parameters $\theta^e$, policy parameters $\phi^e$, and experience replay buffer $\mathcal{D}^e$. Similar to the exploring agent, every chief agent also has an independent environment $\varepsilon^c$. At the beginning of training, each of the exploring agents will first learn the policy independently. Then, each of them will select the best policy it learned, called local optimal policy and denoted by $\pi_{\phi^-}$, and upload it to his chief agent. $\pi_{\phi^-}$ is also called candidate policy in this paper since all the local optimal policies are considered as the candidate for the optimal policy. Next, the chief agent will evaluate the local optimal policies according to their performance and novelty. Then it will upload the policies and their corresponding performance and novelty to the parameter set, which is used to store all the local optimal parameters, and select the current optimal policy among all the policies from the parameter set. After that, the exploring agents will learn their policies with the guidance of the current optimal policy. The chief agents can balance exploration and exploitation dynamically through ascribing different weights to performance and novelty when evaluating a policy, thus guide the whole population for better search in the parameter space of the policy. Detailed introduction can be found in  following subsection. The pseudocode of PNS can be found in Sec. \ref{Pse}.

\subsection{Exploring agents:}
\label{methods_ea}
In PNS, the exploring agent has two roles. The first is to search for the optimal policy. In order to explore the entire parameter space effectively, a population-based searching mechanism is adopted, in which the exploring agent will search the policy under the guidance of the current optimal policy. The update function of the policy parameters $\phi_t^i$ of the $i$-th agent at time step $t$ is
\begin{equation}
    \phi_{t+1}^i \leftarrow  \phi_{t}^i - \alpha\cdot \nabla_{\phi_{t}^i}L(\phi_{t}^i) + c\cdot \gamma \cdot(\phi^* - \phi_{t}^i)\cdot \epsilon_i
    \label{equ_1}
\end{equation}
where $\phi^*$ is the current optimal policy parameters, $L(\phi_{t}^i)$ is the loss function for the Q-function, $\alpha$ is the learning rate, $\gamma$ is random values from 0 to 1, $c$ is the weight of the current optimal policy parameter, which is a hyper-parameter, and $\epsilon_i$ is the regularization term. In Eq.\ref{equ_1}, the third term, $c\cdot \gamma \cdot(\phi^* - \phi_{t}^i)\cdot \epsilon_i$, enables the exploring agents to learn the optimal policy in collaboration, that is, to use the policy information provided by the population to improve exploration, which can help exploring agents avoid local optima and thus improve the performance. The update function can also be formulated by gradient descent for minimizing an augmented loss function $\widehat{L}(\phi^i_t) = L(\phi_t^i)+ \frac{c\cdot\gamma\cdot \epsilon_i}{2\cdot\alpha}\cdot||\phi^*-\phi_t^i||^2_F$.

The other role of the exploring agents is to select the candidate policy $\pi_{\phi^-}$ for the optimal policy. During training, each exploring agent will compare the performance of the current execution policy and the current optimal policy obtained from its chief agent. If the loss of the former is smaller than that of the latter multiplied by a loss coefficient $\tau$,  it will be regarded as a candidate of the optimal policy  $\pi_{\phi^-}$ (\ie, local optimal policy), its parameters and the corresponding Q-value will be uploaded to the chief agent. After that, the chief agent will evaluate the uploaded local optimal policy $\pi_{\phi^-}$ in terms of its performance and novelty.

\subsection{Chief agents}
\label{methods_ca}
In PNS, the role of the chief agents is to find the optimal policy from all local optimal policies and share the current optimal policy with their exploring agents. To obtain a more effective optimal policy and guide future exploration of the entire population, the chief agent evaluates the performance of the candidate policy first, and then its novelty. The cumulative reward $R(\pi_{\phi^-})$  the chief agent gets after executing the candidate policy $\pi_{\phi^-}$ is used to evaluate its performance. After that, the novelty of this policy is computed. The novelty evaluation draws on the approach of novelty search (NS). To evaluate the novelty, we need a domain-dependent behavior characterization (BC) $b(\pi_\phi)$ to describe the behavior of a policy $\pi_\phi$, a cumulative reward $R(\pi_\phi)$ computed before to describe the performance of a policy $\pi$, and a parameter set $\mathcal{A}$ , of which elements are $\{b(\pi_{\phi}), R(\pi_{\phi}),\phi\}$, to store all the policies and their corresponding BC and performance. During the novelty evaluation of each local optimal policy $\pi_{\phi^-}$, the chief agent will first compare its performance. If the performance $R(\pi_{\phi^-})$ is smaller than the mean performance of all the elements in set $\mathcal{A}$, this policy will be abandoned, and the chief agent will move on to evaluate the next local optimal policy. Otherwise, the evaluation continues, and the parameter set $\mathcal{A}$ is updated using the following equation:
\begin{equation}
    \begin{split}
    \mathcal{A} \leftarrow &\{  \{b(\pi_{{\phi_i}}), R(\pi_{{\phi_i}}),\phi_i  \} | R(\pi_{\phi_i})-R(\pi_{\phi^-})>r,\\
    &\{b(\pi_{{\phi_i}}), R(\pi_{{\phi_i}}),\phi_i  \}\in \mathcal{A}\} 
    \end{split}
    \label{equ_2}
\end{equation}
where $r$ is a parameter with a negative value. This step removes all the outdated policies whose performance $R(\pi_{\phi})$ is much smaller than the performance of the current local optimal policy $\pi_{\phi^-}$. Next, the local optimal policy $\pi_{\phi^-}$ will be assigned a corresponding $b(\pi_{\phi^-})$ (see Sec. \ref{methods_bc}), and $\{b(\pi_{\phi^-}), R(\pi_{\phi^-}),\phi^-  \}$ is added to the parameter set $\mathcal{A}$. Finally, the novelty $N(\pi_{{\phi}^-})$ of policy $\pi_{\phi^-}$  is computed by first selecting the k-nearest neighbors of $b(\pi_{{\phi}^-})$ from $\mathcal{A}$ and then computing the average distance between them:
\begin{equation}
    N(\pi_{{\phi}^-}) = \frac{1}{|S|}\sum_{j\in S} ||b(\pi_{{\phi}^-}) - b(\pi_{\phi_j})||_2
     \label{equ_3}
\end{equation}
\begin{equation}
    S = kNN(b(\pi_{{\phi}^-}),A) =\{b(\pi_{\phi_1}),b(\pi_{\phi_2}),...b(\pi_{\phi_k}) \}
    \label{equ_4}
\end{equation}
In Eq.\ref{equ_3}, the distance between BCs is calculated with an L2 norm, but any distance function can be substituted. After the novelty of the candidate policy $\pi_{\phi^-}$ is evaluated and stored, chief agents will select the policy with the highest novelty in set $\mathcal{A}$ as the current optimal policy $\pi_{\phi*}$. Since Eq.\ref{equ_2} only keeps the elements with relatively high reward in set $\mathcal{A}$, the chief agents will eventually select a policy that takes both novelty and high performance into account.

In Eq.\ref{equ_2}, in addition to controlling the size of set $\mathcal{A}$, the parameter $r$ determines the preference of chief agents between performance and novelty. A smaller parameter $r$ means that the chief agent puts more emphasis on novelty. During training, with the change of the policy novelty and performance, and of the preference of chief agents between them, the chief agents can balance exploration and exploitation dynamically, guide the direction for future exploration of the entire population, and thus lead to more effective training.

\subsection{Behavior Characterization (BC)}
\label{methods_bc}
\begin{figure*}
  \centering
  \includegraphics[scale=0.08]{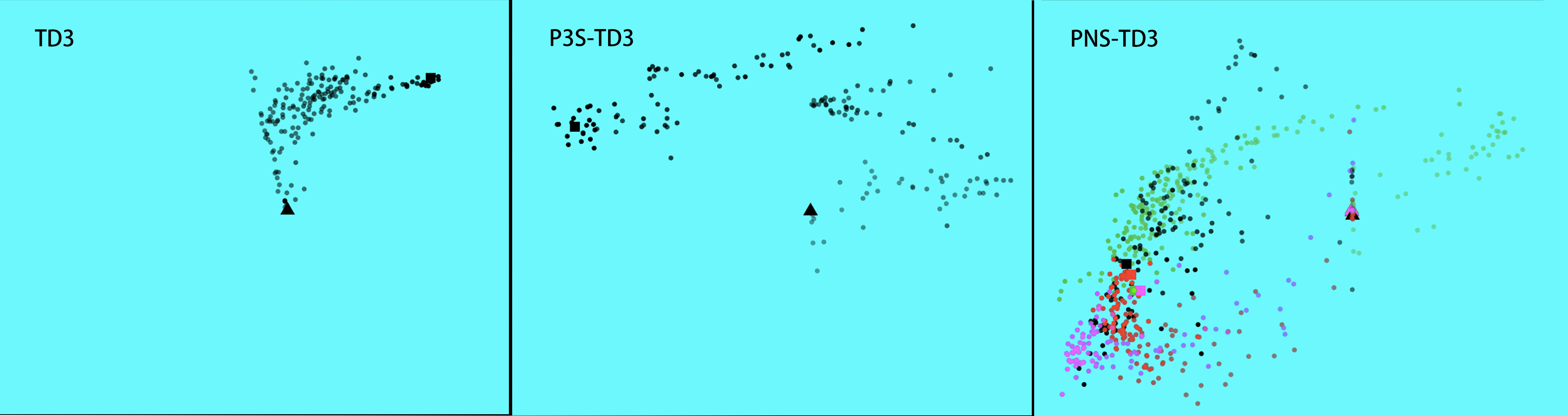}
  \caption{Position of each policy in 2D humanoid locomotion experiment. This figure shows the update of each agent's policy during an entire training process. TD3 gets stuck in the local optimum while P3S-TD3 and PNS-TD3 find a better solution. However, P3S-TD3 still fails to find the optimal position and achieves a reward of 1480, while PNS-TD3 finds the best position and achieves a reward of 2200 (the highest reward in this 2D world).}
  \label{figure_2}
  \vspace{-0.4cm}
\end{figure*}
Another problem that needs to be solved during implementation is the choice of BC. NS requires a domain-dependent BC for each policy to evaluate the novelty. For example, for the Humanoid problem, BC can be the final  $\{x,y\}$ location~\cite{conti2018improving,lehman2011evolving}. In our scheme, in order to make it easier to use, we need to find a more generic BC. Gomez~\cite{gomez2009sustaining} uses action history as generic BC, while Doncieux \etal~\cite{doncieux2013behavioral} build BC based on state counts. Meyerson \etal~\cite{meyerson2016learning} learn BC through a learning method. Although using a more sophisticated BC will improve the performance, it may be hard to construct and compute. In this paper, we develop a standard process to calculate BC. Given a policy, we will let the chief agent explore $m$ timesteps from a random state and record the final state of the agent. This process is repeated for $p$ times. Assuming that the state space of a particular environment is a vector of length $v$, then a matrix of $p\times v$ is obtained. For each column of the matrix, the mean and variance are computed to obtain two vectors of length $v$. After stacking these two vectors into one, we get a vector of length $2\times v$, which can be served as a generic informative BC. For example, for the HalfCheetah problem whose state space is a vector of length $17$. Given a policy, we test it for $p=20$ episodes, and each episode includes executing the policy for $m=50$ timesteps. Then the BC of this policy can be expressed as a vector of length $34$. Due to the use of sub-population and chief agents, this process does not affect the speed of the exploring agents.

\section{Experiments}
To evaluate the effectiveness of PNS, we apply it to the TD3~\cite{fujimoto2018addressing} algorithm and evaluate its performance on the suite of MuJoCo environments~\cite{todorov2012mujoco} and Fetch environments~\cite{brockman2016openai} (simulated robotic benchmarks). 
We also use UR5 robots for real-world experiments. In addition, to prove its efficiency in avoiding local minima and learning from sparse reward signals, we follow Conti \etal~\cite{conti2018improving} and Zheng \etal~\cite{zheng2018learning} and evaluate PNS on two modified simulated benchmarks. Please see the appendix for the ablation study, experiments on PPO, experiment setup, and hyper-parameters.
%Ablation study can be found in Sec. \ref{Abl}. More experimental results on PPO can be found in Sec. \ref{ppo}. The hyper-parameters can be found in Sec. \ref{hyp}.

\subsection{Local optima avoidance and directed exploration} 
\label{e_4.1}
One of the main contributions of this paper is that the introduction of PNS in deep RL can promote direct exploration of the RL algorithm and help the agents avoid local minima. To demonstrate this contribution, we will first analyze its impact on the exploration of parameter space of the policy.

Similar to Conti \etal~\cite{conti2018improving}, we build a 2D humanoid locomotion environment, in which the reward function is only relevant to the agents’ final $\{x,y\}$ position and the highest reward is set to be 2200. During training, the agent needs to learn to walk efficiently and then find a position with the highest reward. To increase the difficulty of training, we also design lots of deceptive rewards, local minima, and wide regions of constant reward in this 2D world. For this problem, the BC is the final $\{x,y\}$ location. 

To demonstrate the effectiveness of PNS for local optima avoidance and directed exploration, we use TD3, P3S-TD3~\cite{jung2020population}, and PNS-TD3 to train agents in this environment until convergence. Then we visualize the policies explored by each agent of these algorithms and compare the final rewards. P3S-TD3 is currently the highest-performing population-based scheme. For PNS-TD3, we randomly select 4 exploring agents for visualization. For TD3, we test 24 times and select the agent with the highest performance. For P3S-TD3, since it will find the best learner during training and then search the spread area with radius $d$ around the best learner for a better policy, we only visualize the policy of the current best learner. The update of each agent's policy during the training process is shown in Fig. \ref{figure_2}. Different color represents the policy of different agents, with darker shading for later generations. The triangle represents the starting point of an agent and the square represents the final policy of the agent. As can be observed from Fig. \ref{figure_2}, TD3 converges to a point only after exploring a small area of the parameter space, while P3S-TD3 and PNS-TD3 explore a larger area. 

If we take a close look at the trajectory of PNS-TD3 agents, it can be observed that at the beginning of training, one agent (visualized as green points) moves toward the position where TD3 converges, proving that it is indeed a local optimum. However, after weighing the performance and novelty, the optimal policy selected by the chief agents helps this green agent escape local optima. P3S-TD3 also escapes local optima where TD3 gets stuck. It can be observed that both P3S and PNS can encourage exploration of the algorithm and help the agent avoid local optima. 

In this experiment, P3S-TD3 and PNS-TD3 converge to a different position and achieve different final rewards. TD3 only achieves a reward of 849 as the final reward, P3S-TD3 achieves a reward of 1480, and PNS-TD3 achieves a reward of 2200, which is also the highest reward an agent can achieve in this 2D world. From the final reward we can observe that PNS has a better exploration efficiency and ability to address the problem of local optima than P3S.

During the early period of training, which is represented as the points with shallower shading, the novelty of each policy is high while the performance is relatively low. At this period, the agents will put more emphasis on exploring new policies and they tend to explore different parts of the environment as shown in the training process of PNS in Fig. \ref{figure_2}. As training goes on, the chief agents will balance the trade-off between exploration and exploitation according to the performance and the novelty of each policy. If an agent gets stuck in local optima, the performance will no longer be enhanced and the policy of the agent is no longer updated, thus the novelty of this policy that leads to local optima will decrease. The chief agents will encourage the agents to escape this area, and explore those areas with relatively higher novelty. During the late period of training, which is represented as darker shading points, the novelty of all policies decreases as the entire space is explored. The population distribution gradually moves closer to the position where the optimal policy can be obtained. Multiple agents explore near multiple possible optimal positions, allocating more computational resources for positions that are more likely to achieve high rewards. It dynamically balances the trade-off between exploration and exploitation during different training periods. 

\subsection{Learning from sparse reward signals}
\label{e_4.2}
\begin{table}
    \caption{Performance of PNS-TD3 and TD3 in real robot experiments. The results show the Eular distance, in centimeter, between the position of the target and the end effector. Our setup only requires four robotic arms and four cameras. TD3 cannot learn anything from sparse rewards, but PNS-TD3  enables robotic agents to learn skills efficiently from sparse signals.}
    \label{tab:1}
    \small 
    \centering
    \begin{tabular}{cccccc}
    \toprule
    Training epoch & 100 &200&300&400&500\\
    \midrule
    PNS-TD3&35.22&10.18&4.20&4.13&4.29\\
    TD3&42.45&46.21&40.83&38.14&41.52\\
    \bottomrule
    \end{tabular}
    \vspace{-0.4cm}
\end{table}
\begin{figure}
  \centering
  \includegraphics[scale=0.25]{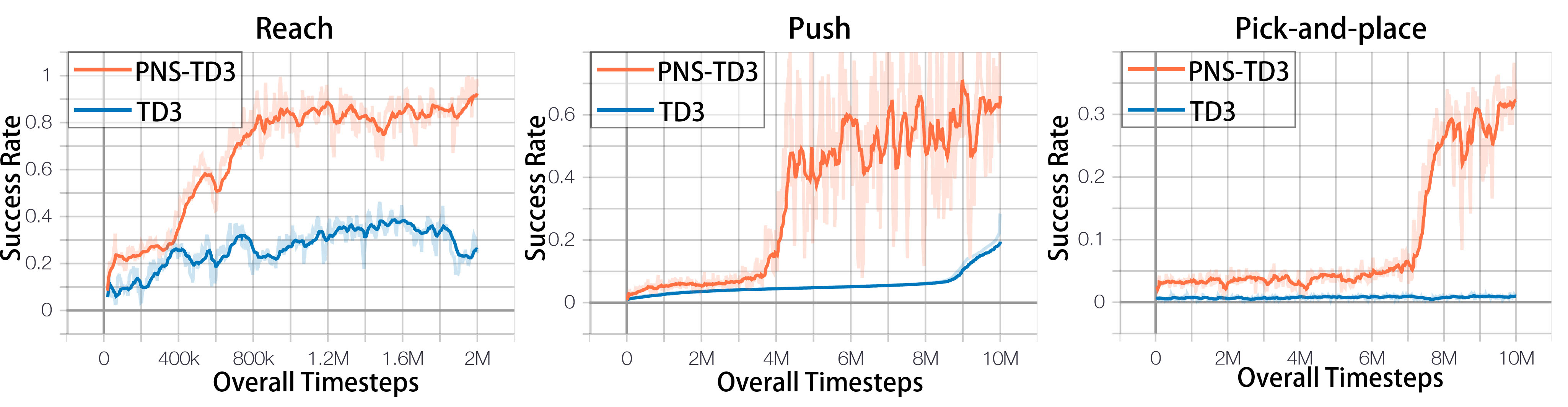}
  \caption{Success rate of PNS-TD3 and TD3 in Fetch environments.}
  \label{figure_r}
\end{figure}
Learning from sparse reward functions is still a problem when training RL agents for robotic tasks. In this section, we demonstrate that PNS can enable robotics agents to learn efficiently from sparse reward signals (see Fig. \ref{teaser}). For simulation experiments, we consider the reach, push, and pick-and-place tasks from the OpenAI Gym Fetch environment. The results are provided in Fig. \ref{figure_r}. We use UR5 robot for real-world experiments and consider the task of reaching a given point. We evaluate the results every 100 training epochs. During the evaluation,  we fix a marker on the end effector to compute the distance between the final position of the end effector and the target. The results can be found in Table \ref{tab:1}. In all experiments, we use images as input and only provide success or not as the reward signals. From the results, we can found that PNS enables the baseline method to learn efficiently from sparse reward signals.

\subsection{Learning from delayed reward signals}
\label{e_4.3}
\begin{figure}
  \centering
  \includegraphics[scale=0.23]{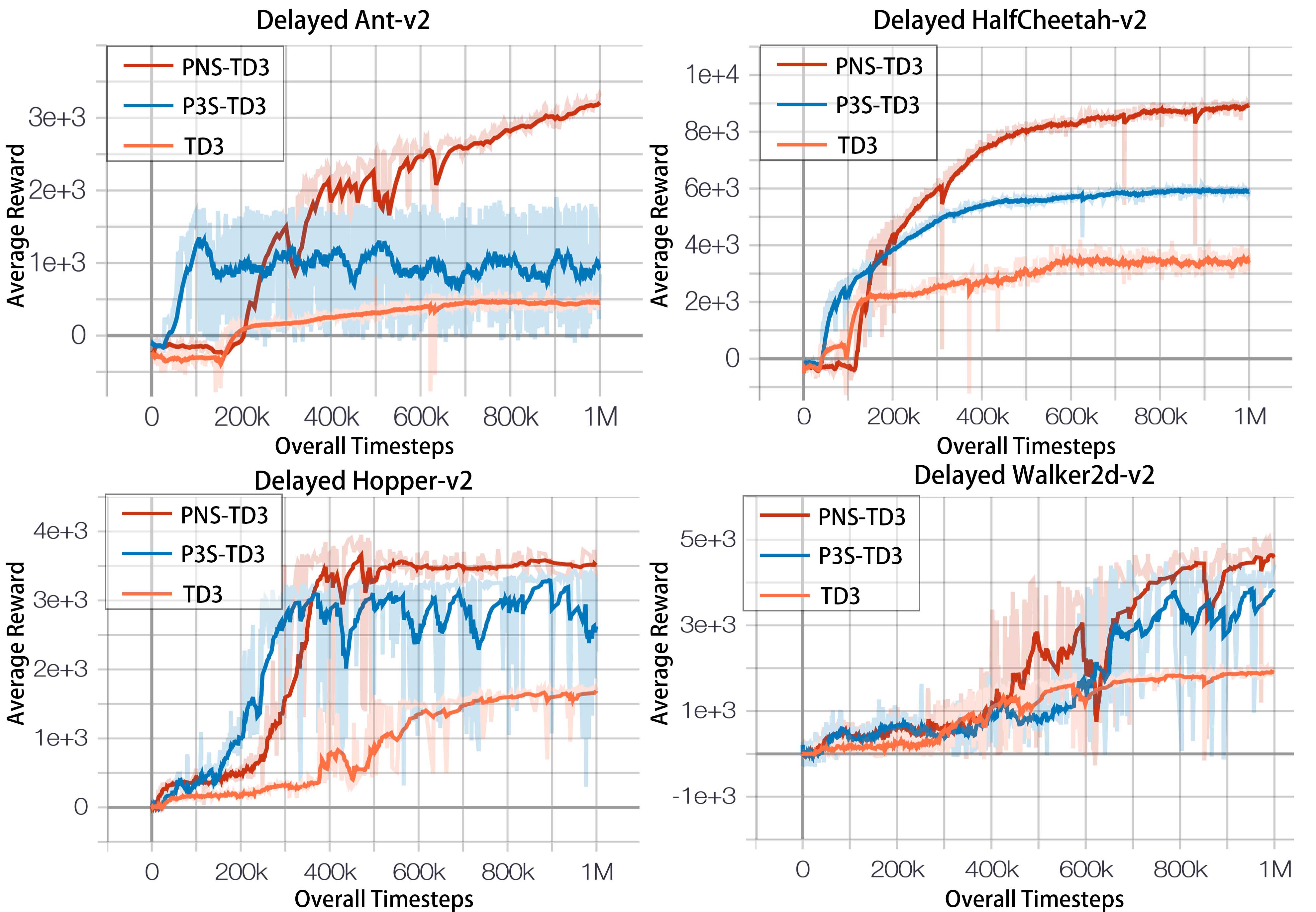}
  \caption{Performance of different population-based methods in Delayed MuJoCo environments.}
  \label{figure_5}
\end{figure}
To demonstrate the efficiency of PNS to learn from delayed reward signals, we also evaluate the performance of PNS in delayed reward environments (\ie,  Delayed MuJoCo environments used in Zheng \etal~\cite{zheng2018learning}). Delayed MuJoCo environments are reward-sparsified versions of MuJoCo environments. They only give non-zero rewards periodically with frequency $f$ or at the end of episodes, which require more exploration to obtain a good policy. We evaluate the performance of PNS-TD3 against P3S-TD3 (the SOTA method), and the learning curves are shown in Fig. \ref{figure_5}. It can be observed that in these hard exploration environments, PNS-TD3 takes off a bit slower than P3S-TD3, but it outperforms P3S-TD3 a lot. Especially in the problem of delayed Ant where P3S-TD3 achieves an average reward of 1740 but PNS-TD3 achieves an average reward of 3395. The policy PNS learned is also more stable than the policy P3S-TD3 learned in the problem of delayed Ant and delayed Hopper. Empirically, PNS is more efficient than the SOTA population-based method in delayed reward settings.

\begin{table*}
    \caption{Performance of different population-based methods and distributed algorithms in MuJoCo environments. We select the best results among 24 agents.}
    \label{bas}
    \small 
    \centering
    \begin{tabular}{cccccccc}
    \toprule
     & PNS-TD3&TD3&P3S-TD3&CEM-TD3&PBT-TD3&IMPACT&IMPALA\\
    \midrule
    Ant&\textbf{5937}&4631&5391&4302&5120&5212&1784\\
    HalfCheetah&\textbf{12820}&10035&11530&10891&10943&11023&2130\\
    Hopper&\textbf{3891}&3681&3741&3653&3680&3458&895\\
    Walker2d&\textbf{5504}&4693&4896&4793&5014&4201&1103\\
    \bottomrule
    \end{tabular}
\end{table*}

\begin{figure*}
  \centering
  \includegraphics[scale=0.23]{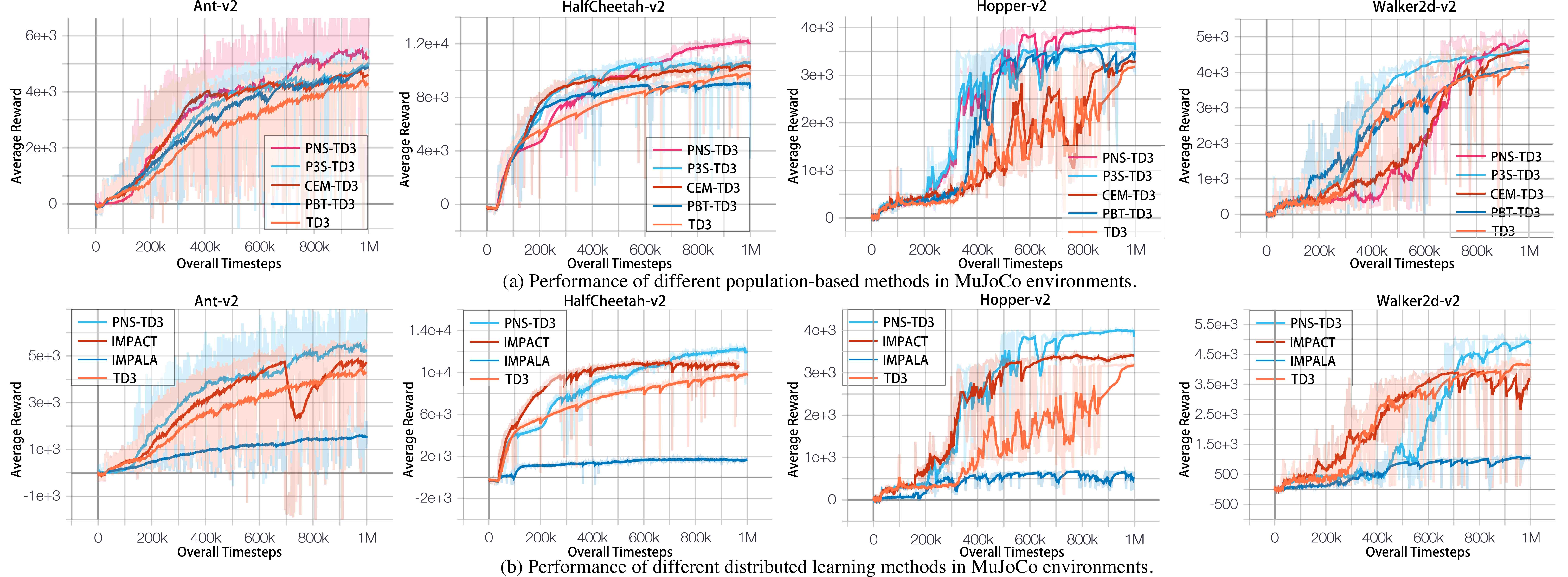}
  \caption{Performance of different population-based methods (a) and different distributed algorithms (b) in MuJoCo environments.}
  \label{figure_4}
  \vspace{-0.4cm}
\end{figure*}
\begin{figure}
  \centering
  \includegraphics[scale=0.06]{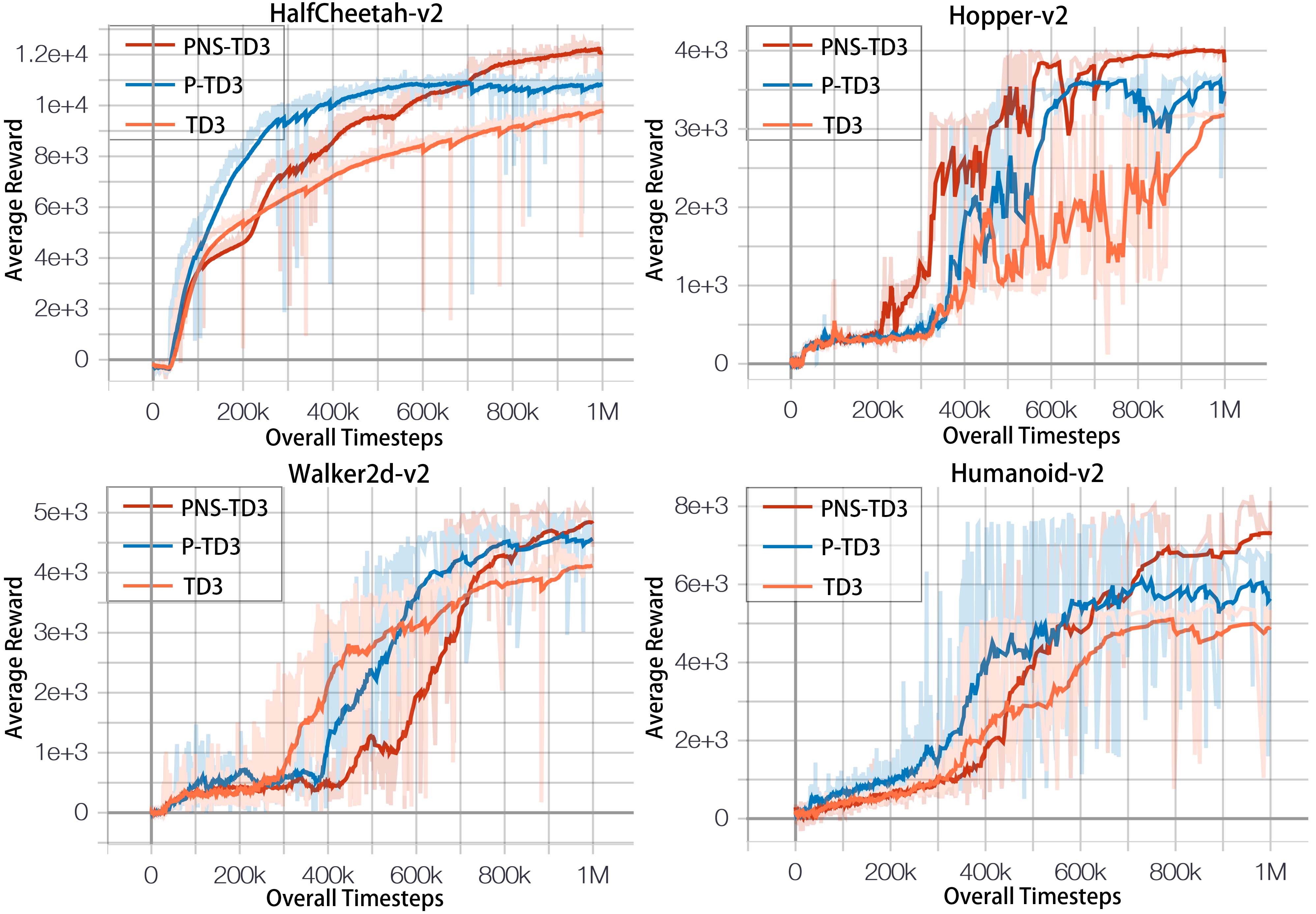}
  \caption{Performance of PNS-TD3(red), P-TD3(blue) and TD3(orange) in MuJoCo environments.}
  \label{figure_3}
\end{figure}

\subsection{Comparison with baselines}
\label{e_4.2}
In the previous subsection, we analyze the improvement of exploration. Next, we will demonstrate the improved performance of PNS. We first compare PNS-TD3 with baseline algorithms in this subsection, then we compare it with other SOTA distributed algorithms in the following subsection. 

To demonstrate the effectiveness of using NS in deep RL, we modify the proposed PNS scheme. We remove the novelty evaluation part from the chief agents (\ie, now the chief agents only evaluate the policies by performance), and keep the rest of the structure and hyper-parameters to obtain the P-TD3 algorithm and use it as one baseline algorithm. We also use TD3 as another baseline algorithm.

Fig. \ref{figure_3} shows the learning curves for several continuous control domains. We first compare PNS-TD3 with TD3. It can be observed that the PNS-TD3 outperforms the baseline algorithms in terms of both the speed of convergence and the final performance. Then we compare PNS-TD3 with P-TD3. We can find that PNS-TD3 yields improvement in the problem of HalfCheetah, Hopper, and Humanoid. However, it doesn’t yield a significant improvement in the problem of Walker2d compared with P-TD3. We believe that there are mainly two reasons: One is that population-based training itself can also promote the exploration, so the additional improvement yielded by introducing NS can be observed in some problems but may not be obvious for other problems. The second reason is that the generic BC used in PNS may not be ideal in some MuJoCo problems. How to choose BC is still a question worth studying~\cite{pugh2015confronting,meyerson2016learning}. However, our work aims to study the impact of introducing NS to deep RL algorithms, and the existing conclusions are sufficient to prove its efficiency, so we can leave the task of finding more generic and informative BC for future work.
%that the introduction of NS has made improvement compared to the original algorithm and population-based training, so we can leave the task of finding more generic and informative BC for future work.

We can also observe that in HalfCheetah, Walk2d, and Humanoid, PNS-TD3 learns slower than P-TD3 in the early stage. It’s because in this stage PNS prefers policy with higher novelty but maybe low-rewards but P-TD3 and other algorithms will always search for high-rewards policy. It is another example demonstrates that PNS can dynamically balance the trade-off between exploration and exploitation. 

\subsection{Comparison with other parallel learning schemes}
Then, we compare the PNS scheme with other population-based schemes (\ie, P3S-TD3~\cite{jung2020population}, CEM-TD3~\cite{pourchot2018cem}, and PBT-TD3~\cite{jaderberg2017population}), and other distributed algorithms (\ie, IMPACT~\cite{luo2019impact}, IMPALA~\cite{espeholt2018impala}). The learning curves are shown in Fig. \ref{figure_4} and the best rewards of each algorithm can be found in Table \ref{bas}. All the algorithms use 24 agents. The performance of IMPALA is surprisingly bad in our experiments. However, the reported performance is the same as the results shown in Luo \etal~\cite{luo2019impact}. The reason for the bad performance could be that vanilla IMPALA is not very effective in continuous domains. It can be observed that PNS-TD3 outperforms most of the SOTA population-based and distributed algorithms.  

As mentioned before, PNS prefers policy with higher novelty but maybe low-rewards during the early period of training, which means PNS-TD3 may learn slower than P3S-TD3 in the early stage. Therefore, in some cases such as the problem of HalfCheetah, P3S-TD3 learns faster in the early stage, but PNS-TD3 can find a better policy. 

\section{Discussion and Conclusion}

In this paper, we introduce a Population-guided Novelty Search (PNS) method for deep RL algorithms. The scheme is composed of multiple sub-populations, each of which has one chief agent and several exploring agents. 
%The role of the chief agent is to evaluate the performance and novelty of the policies learned by the exploring agents and to share the optimal policy with all sub-populations. The role of exploring agents is to update their policies in collaboration with the guidance of the optimal policy and, simultaneously, upload their policies to the chief agent. 
Compared to other population-based parallel schemes, PNS uses chief agents and novelty search to encourage the population distribution to move toward areas of policy space with high performance and high novelty. Our experiments confirm that this method can promote directed exploration of RL algorithms and enable better policy search, and the collaboration among agents and sub-populations can help the agents avoid local optima. In addition, we demonstrate that PNS can learn efficiently from sparse reward and delayed reward signals in robot manipulation and other continuous control tasks.
\section*{APPENDIX}
\subsection{Pseudocode for PNS}
\label{Pse}
\vspace{-0.4cm}
\begin{algorithm}[htbp]
\caption{Pseudocode for each chief agent}
\label{pse}
\small
\begin{algorithmic}[1] 
\Require $r$: preference between performance and novelty, $m$: time steps for calculating BC, $p$: maximum test times for calculating BC, $T$: maximum time steps
\State Initialize candidate policy $\pi$ with parameters $\phi^-$
\State Initialize cumulative reward $R(\pi_{\phi^-}) \leftarrow 0$
\State Get initial state $s_0$
\While{$t<T$}
    \State Take action $a_t=\pi_{\phi^-} (s_t)+\epsilon$ and observe reward $r_t$ and next state $s_{t+1}$
    \State Accumulate reward $R(\pi_{\phi^-}) \leftarrow R(\pi_{\phi^-}) + r_t$
    \State $t\leftarrow t+1$
\EndWhile
\State Compute mean performance $R_{mean}$ of all the elements in $\mathcal{A}$
\If{$R(\pi_{\phi^-}) > R_{mean}$}
\State Update the parameter set:
\begin{align*}
    \begin{split}
    \mathcal{A} \leftarrow &\{  \{b(\pi_{{\phi_i}}), R(\pi_{{\phi_i}}),\phi_i  \} | R(\pi_{\phi_i})-R(\pi_{\phi^-})>r,\\
    &\{b(\pi_{{\phi_i}}), R(\pi_{{\phi_i}}),\phi_i  \}\in \mathcal{A}\} 
    \end{split}
\end{align*}
\While{$k<p$}
\State Take a random state as the starting state $s^k_0$
\While{$j<m$}
\State Take action $a^k_j=\pi_{\phi^-} (s^k_j)+\epsilon$ and observe next state $s^k_{j+1}$
\State $j\leftarrow j+1$
\EndWhile
\State $k\leftarrow k+1$
\EndWhile
\State Compute the mean and variance of $s^k_m$ and get $b(\pi_{\phi^-})$
\State Store $\{b(\pi_{\phi^-}),R(\pi_{\phi^-}),\phi^-\}$ to the parameter set $\mathcal{A}$
\State Compute the novelty:
\begin{align*}
    N(\pi_{{\phi}^-}) = \frac{1}{|S|}&\sum_{j\in S} ||b(\pi_{{\phi}^-}) - b(\pi_{\phi_j})||_2 \\
    S = kNN(b(\pi_{{\phi}^-}),A) &=\{b(\pi_{\phi_1}),b(\pi_{\phi_2}),...b(\pi_{\phi_k}) \}
\end{align*}
\State Store $N(\pi_{{\phi}^-})$ to a set $\mathcal{B}$ \algorithmiccomment{In our source code, set $\mathcal{B}$ is absorbed into set $\mathcal{A}$}
\EndIf
\State Select the policy with the highest novelty in set $\mathcal{A}$ as the current optimal policy $\pi_\phi^*$
\newpage
\end{algorithmic}
\end{algorithm}
\vspace{-0.4cm}

\subsection{Ablation study on the number of sub-populations}
\label{Abl}
\begin{figure}
  \centering
  \includegraphics[scale=0.23]{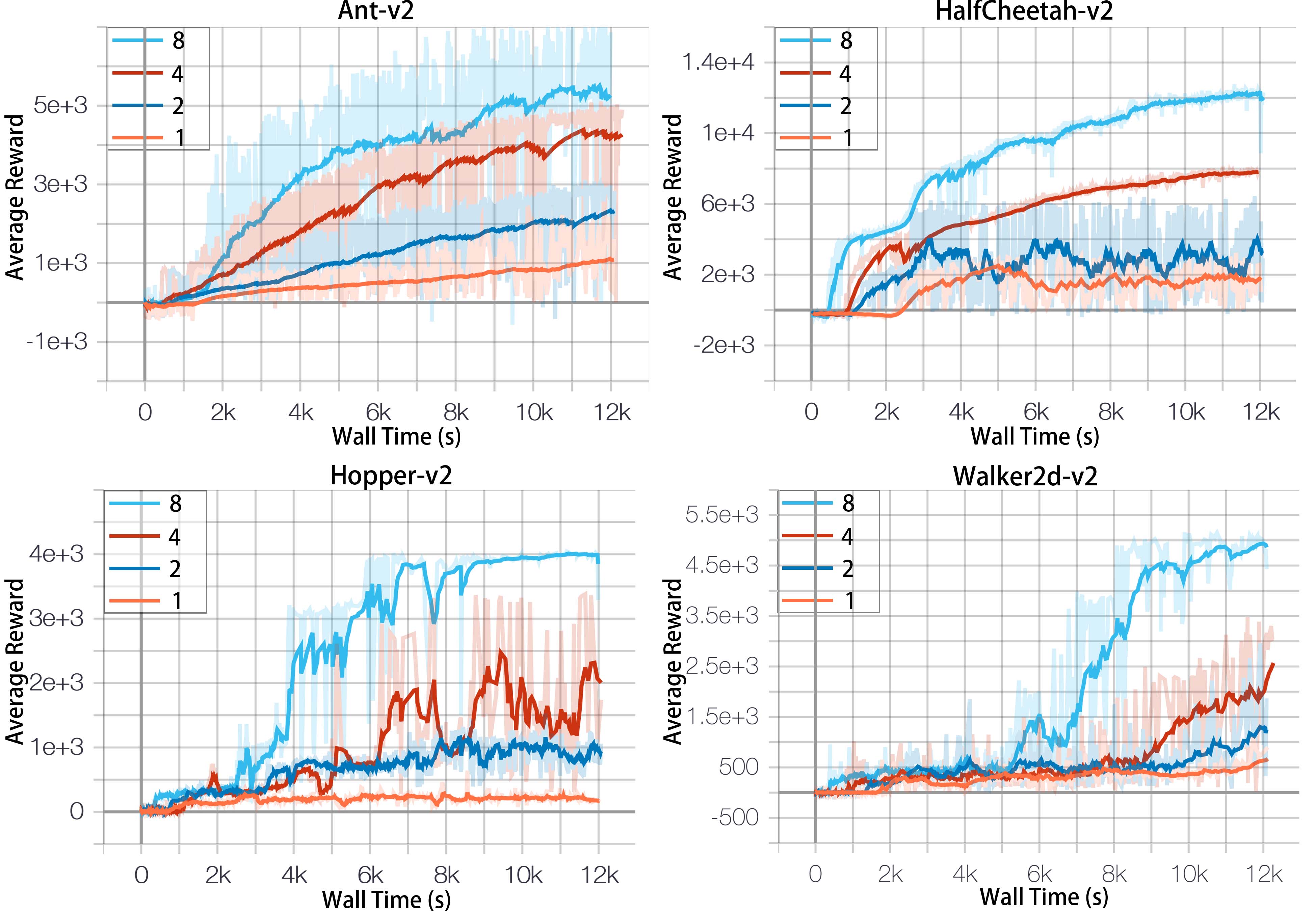}
  \caption{Performance of PNS-TD3 with a fixed total number of exploring agents (=24) but a different number (n=1,2,4,8) of sub-populations in terms of wall clock time.}
  \vspace{-0.4cm}
  \label{figure_6}
\end{figure}
In this subsection, we will study the impact of introducing sub-population on the performance of each algorithm. Two of the main challenges when using NS and synchronous framework are that:(1) NS occupies a lot of computer resources and (2) the transmit of weight is time-wasting and the global barrier is placed for each iteration~\cite{dean2012large,li2014communication}. It takes $49.9\%-83.2\%$ of the execution time of each iteration for the
network communication in widely-used synchronous distributed RL schemes~\cite{li2019accelerating}. Our study shows that the average time each agent takes to update for one step in vanilla TD3 and PNS-TD3 with a single population of 24 agents is 0.0139s and 0.0374s, respectively. To minimize the blocking overhead, asynchronous training is used~\cite{ho2013more,nair2015massively}. However, it suffers from the staleness of local weight in training works, which slows down training convergence~\cite{ho2013more} and increases the number of training iterations. This problem is more serious in our proposed frame since NS requires up-to-date policies for better performance.
Take PNS into account, we introduce sub-population and divide the agents into small groups. Then we use chief agents to perform NS and guide the exploration of this small group. Fig. \ref{figure_6} shows the performance of PNS-TD3 with a fixed total number of exploring agents but a different number of sub-populations in terms of wall clock time. It takes 0.031s to update for one step when using 2 sub-populations, 0.023s when using 4 sub-populations, and 0.0141s when using 8. It can be observed that using sub-population can save the wall-clock time and avoid sacrificing data efficiency.

\label{ppo}
\begin{figure}
  \centering
  \includegraphics[scale=0.04]{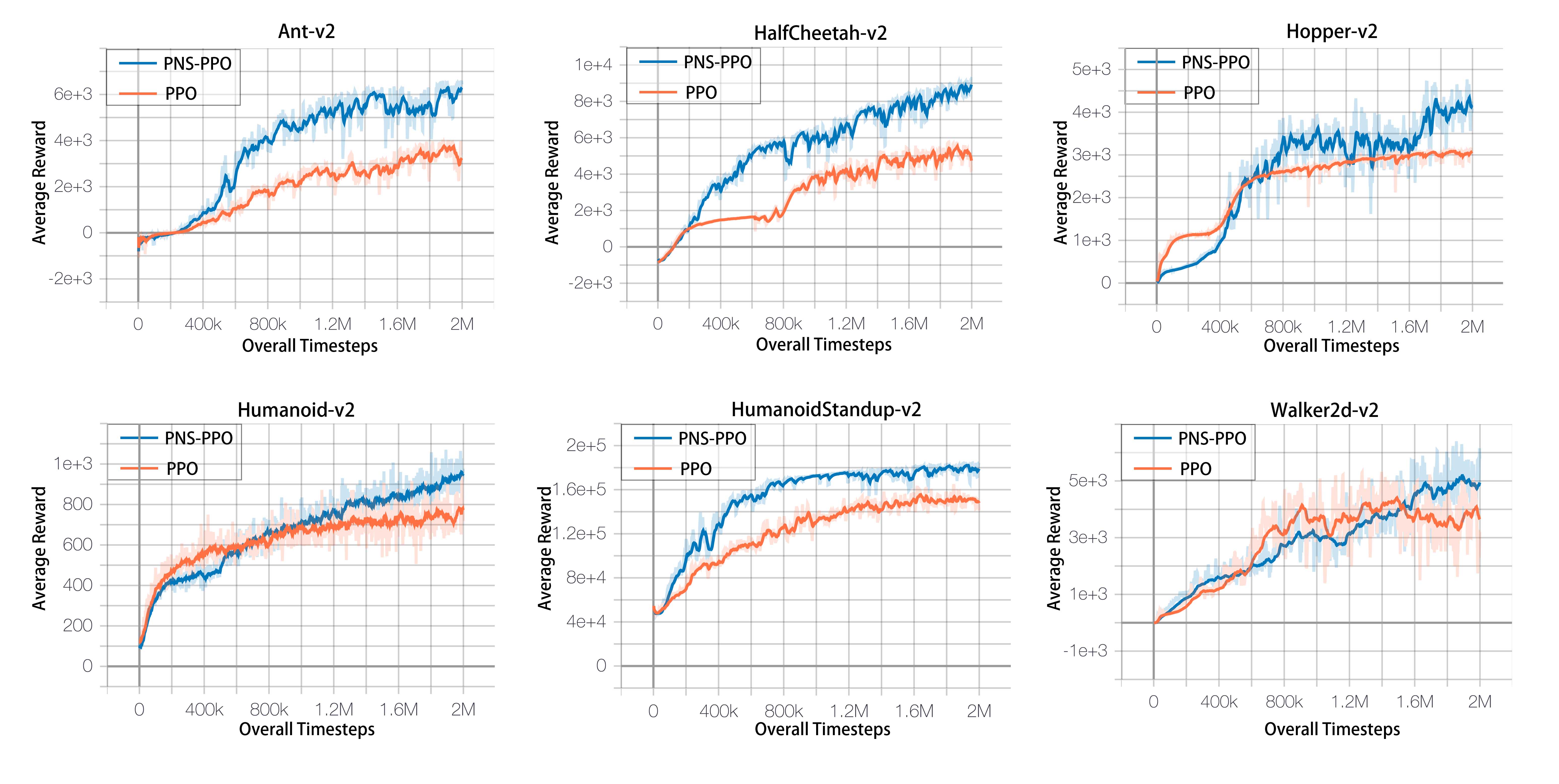}
  \vspace{-0.6cm}
  \caption{Performance for PNS-PPO(blue) and PPO(orange) in MuJoCo.}
  \label{APP}
\end{figure}

\subsection{Real-robot learning setup}
We consider the reaching task of the UR5 robot (Fig.~\ref{teaser}). The target object is a cylinder with 12 cm in diameter. The robot is set to be able to move within a workspace of approximately 55 by 40 cm. The cylinder is randomly set in the workspace and the end effector is initialized above the center of the workspace. To better evaluate the performance, we attach a marker pen to the end effector and cover the workspace with blank paper, which gives us the distance between the final position of the end effector and the target. 

\subsection{Hyper-parameters}
\label{hyp}
The hyper-parameters of all the baseline algorithms and other state-of-the-art schemes adopted in this paper are the same as those in the original papers. Here, only the hyper-parameters of the PNS scheme are provided. On top of the hyper-parameters of the base TD3, we all use $n=8$ sub-populations, each sub-population has $k=3$ exploring agents for PNS-TD3. To update the policy of each exploring agent, the weight $c=3\times 10^{-4}$ is used in PNS-TD3. For BC, $m = 50$ and $p=20$ is used. All experiments use $1$ CPU with $24$ cores and $2$ GPUs. All policy and Q-function networks consist of three fully-connected layers, with non-linear activation ReLU. To be fair, each baseline algorithm uses $24$ different random seeds to test $24$ times in each environment, and the best result is selected for comparison.
%\section*{ACKNOWLEDGMENT}

%%%%%%%%%%%%%%%%%%%%%%%%%%%%%%%%%%%%%%%%%%%%%%%%%%%%%%%%%%%%%%%%%%%%%%%%%%%%%%%%

\bibliographystyle{ieeetr}
\bibliography{ref}

\begin{thebibliography}{10}

\bibitem{osband2016deep}
I.~Osband, C.~Blundell, A.~Pritzel, and B.~Van~Roy, ``Deep exploration via
  bootstrapped dqn,'' in {\em Advances in neural information processing
  systems}, pp.~4026--4034, 2016.

\bibitem{zhu2019dexterous}
H.~Zhu, A.~Gupta, A.~Rajeswaran, S.~Levine, and V.~Kumar, ``Dexterous
  manipulation with deep reinforcement learning: Efficient, general, and
  low-cost,'' in {\em 2019 International Conference on Robotics and Automation
  (ICRA)}, pp.~3651--3657, IEEE, 2019.

\bibitem{gu2017deep}
S.~Gu, E.~Holly, T.~Lillicrap, and S.~Levine, ``Deep reinforcement learning for
  robotic manipulation with asynchronous off-policy updates,'' in {\em 2017
  IEEE international conference on robotics and automation (ICRA)},
  pp.~3389--3396, IEEE, 2017.

\bibitem{andrychowicz2017hindsight}
M.~Andrychowicz, F.~Wolski, A.~Ray, J.~Schneider, R.~Fong, P.~Welinder,
  B.~McGrew, J.~Tobin, P.~Abbeel, and W.~Zaremba, ``Hindsight experience
  replay,'' {\em arXiv preprint arXiv:1707.01495}, 2017.

\bibitem{zhao2019curiosity}
R.~Zhao and V.~Tresp, ``Curiosity-driven experience prioritization via density
  estimation,'' {\em arXiv preprint arXiv:1902.08039}, 2019.

\bibitem{riedmiller2018learning}
M.~Riedmiller, R.~Hafner, T.~Lampe, M.~Neunert, J.~Degrave, T.~Wiele, V.~Mnih,
  N.~Heess, and J.~T. Springenberg, ``Learning by playing solving sparse reward
  tasks from scratch,'' in {\em International Conference on Machine Learning},
  pp.~4344--4353, PMLR, 2018.

\bibitem{inoue2017deep}
T.~Inoue, G.~De~Magistris, A.~Munawar, T.~Yokoya, and R.~Tachibana, ``Deep
  reinforcement learning for high precision assembly tasks,'' in {\em 2017
  IEEE/RSJ International Conference on Intelligent Robots and Systems (IROS)},
  pp.~819--825, IEEE, 2017.

\bibitem{bellemare2016unifying}
M.~Bellemare, S.~Srinivasan, G.~Ostrovski, T.~Schaul, D.~Saxton, and R.~Munos,
  ``Unifying count-based exploration and intrinsic motivation,'' in {\em
  Advances in neural information processing systems}, pp.~1471--1479, 2016.

\bibitem{haber2018learning}
N.~Haber, D.~Mrowca, S.~Wang, L.~F. Fei-Fei, and D.~L. Yamins, ``Learning to
  play with intrinsically-motivated, self-aware agents,'' in {\em Advances in
  Neural Information Processing Systems}, pp.~8388--8399, 2018.

\bibitem{houthooft2016vime}
R.~Houthooft, X.~Chen, Y.~Duan, J.~Schulman, F.~De~Turck, and P.~Abbeel,
  ``Vime: Variational information maximizing exploration,'' in {\em Advances in
  Neural Information Processing Systems}, pp.~1109--1117, 2016.

\bibitem{pathak2017curiosity}
D.~Pathak, P.~Agrawal, A.~A. Efros, and T.~Darrell, ``Curiosity-driven
  exploration by self-supervised prediction,'' in {\em Proceedings of the IEEE
  Conference on Computer Vision and Pattern Recognition Workshops}, pp.~16--17,
  2017.

\bibitem{jaderberg2017population}
M.~Jaderberg, V.~Dalibard, S.~Osindero, W.~M. Czarnecki, J.~Donahue, A.~Razavi,
  O.~Vinyals, T.~Green, I.~Dunning, K.~Simonyan, {\em et~al.}, ``Population
  based training of neural networks,'' {\em arXiv preprint arXiv:1711.09846},
  2017.

\bibitem{jaderberg2019human}
M.~Jaderberg, W.~M. Czarnecki, I.~Dunning, L.~Marris, G.~Lever, A.~G.
  Castaneda, C.~Beattie, N.~C. Rabinowitz, A.~S. Morcos, A.~Ruderman, {\em
  et~al.}, ``Human-level performance in 3d multiplayer games with
  population-based reinforcement learning,'' {\em Science}, vol.~364, no.~6443,
  pp.~859--865, 2019.

\bibitem{jung2020population}
W.~Jung, G.~Park, and Y.~Sung, ``Population-guided parallel policy search for
  reinforcement learning,'' {\em arXiv preprint arXiv:2001.02907}, 2020.

\bibitem{nair2015massively}
A.~Nair, P.~Srinivasan, S.~Blackwell, C.~Alcicek, R.~Fearon, A.~De~Maria,
  V.~Panneershelvam, M.~Suleyman, C.~Beattie, S.~Petersen, {\em et~al.},
  ``Massively parallel methods for deep reinforcement learning,'' {\em arXiv
  preprint arXiv:1507.04296}, 2015.

\bibitem{mnih2016asynchronous}
V.~Mnih, A.~P. Badia, M.~Mirza, A.~Graves, T.~Lillicrap, T.~Harley, D.~Silver,
  and K.~Kavukcuoglu, ``Asynchronous methods for deep reinforcement learning,''
  in {\em International conference on machine learning}, pp.~1928--1937, 2016.

\bibitem{babaeizadeh2016reinforcement}
M.~Babaeizadeh, I.~Frosio, S.~Tyree, J.~Clemons, and J.~Kautz, ``Reinforcement
  learning through asynchronous advantage actor-critic on a gpu,'' {\em arXiv
  preprint arXiv:1611.06256}, 2016.

\bibitem{stooke2018accelerated}
A.~Stooke and P.~Abbeel, ``Accelerated methods for deep reinforcement
  learning,'' {\em arXiv preprint arXiv:1803.02811}, 2018.

\bibitem{horgan2018distributed}
D.~Horgan, J.~Quan, D.~Budden, G.~Barth-Maron, M.~Hessel, H.~Van~Hasselt, and
  D.~Silver, ``Distributed prioritized experience replay,'' {\em arXiv preprint
  arXiv:1803.00933}, 2018.

\bibitem{espeholt2018impala}
L.~Espeholt, H.~Soyer, R.~Munos, K.~Simonyan, V.~Mnih, T.~Ward, Y.~Doron,
  V.~Firoiu, T.~Harley, I.~Dunning, {\em et~al.}, ``Impala: Scalable
  distributed deep-rl with importance weighted actor-learner architectures,''
  {\em arXiv preprint arXiv:1802.01561}, 2018.

\bibitem{luo2019impact}
M.~Luo, J.~Yao, E.~Liang, R.~Liaw, and I.~Stoica, ``Impact: Importance weighted
  asynchronous architectures with clipped target networks,'' {\em arXiv
  preprint arXiv:1912.00167}, 2019.

\bibitem{vinyals2019alphastar}
O.~Vinyals, I.~Babuschkin, J.~Chung, M.~Mathieu, M.~Jaderberg, W.~M. Czarnecki,
  A.~Dudzik, A.~Huang, P.~Georgiev, R.~Powell, {\em et~al.}, ``Alphastar:
  Mastering the real-time strategy game starcraft ii,'' {\em DeepMind blog},
  p.~2, 2019.

\bibitem{arulkumaran2019alphastar}
K.~Arulkumaran, A.~Cully, and J.~Togelius, ``Alphastar: An evolutionary
  computation perspective,'' in {\em Proceedings of the Genetic and
  Evolutionary Computation Conference Companion}, pp.~314--315, 2019.

\bibitem{conti2018improving}
E.~Conti, V.~Madhavan, F.~P. Such, J.~Lehman, K.~Stanley, and J.~Clune,
  ``Improving exploration in evolution strategies for deep reinforcement
  learning via a population of novelty-seeking agents,'' in {\em Advances in
  neural information processing systems}, pp.~5027--5038, 2018.

\bibitem{lehman2011evolving}
J.~Lehman and K.~O. Stanley, ``Evolving a diversity of virtual creatures
  through novelty search and local competition,'' in {\em Proceedings of the
  13th annual conference on Genetic and evolutionary computation},
  pp.~211--218, 2011.

\bibitem{gomez2009sustaining}
F.~J. Gomez, ``Sustaining diversity using behavioral information distance,'' in
  {\em Proceedings of the 11th Annual conference on Genetic and evolutionary
  computation}, pp.~113--120, 2009.

\bibitem{doncieux2013behavioral}
S.~Doncieux and J.-B. Mouret, ``Behavioral diversity with multiple behavioral
  distances,'' in {\em 2013 IEEE Congress on Evolutionary Computation},
  pp.~1427--1434, IEEE, 2013.

\bibitem{meyerson2016learning}
E.~Meyerson, J.~Lehman, and R.~Miikkulainen, ``Learning behavior
  characterizations for novelty search,'' in {\em Proceedings of the Genetic
  and Evolutionary Computation Conference 2016}, pp.~149--156, 2016.

\bibitem{fujimoto2018addressing}
S.~Fujimoto, H.~Van~Hoof, and D.~Meger, ``Addressing function approximation
  error in actor-critic methods,'' {\em arXiv preprint arXiv:1802.09477}, 2018.

\bibitem{todorov2012mujoco}
E.~Todorov, T.~Erez, and Y.~Tassa, ``Mujoco: A physics engine for model-based
  control,'' in {\em 2012 IEEE/RSJ International Conference on Intelligent
  Robots and Systems}, pp.~5026--5033, IEEE, 2012.

\bibitem{brockman2016openai}
G.~Brockman, V.~Cheung, L.~Pettersson, J.~Schneider, J.~Schulman, J.~Tang, and
  W.~Zaremba, ``Openai gym,'' {\em arXiv preprint arXiv:1606.01540}, 2016.

\bibitem{zheng2018learning}
Z.~Zheng, J.~Oh, and S.~Singh, ``On learning intrinsic rewards for policy
  gradient methods,'' in {\em Advances in Neural Information Processing
  Systems}, pp.~4644--4654, 2018.

\bibitem{pugh2015confronting}
J.~K. Pugh, L.~B. Soros, P.~A. Szerlip, and K.~O. Stanley, ``Confronting the
  challenge of quality diversity,'' in {\em Proceedings of the 2015 Annual
  Conference on Genetic and Evolutionary Computation}, pp.~967--974, 2015.

\bibitem{pourchot2018cem}
A.~Pourchot and O.~Sigaud, ``Cem-rl: Combining evolutionary and gradient-based
  methods for policy search,'' {\em arXiv preprint arXiv:1810.01222}, 2018.

\bibitem{dean2012large}
J.~Dean, G.~Corrado, R.~Monga, K.~Chen, M.~Devin, M.~Mao, M.~Ranzato,
  A.~Senior, P.~Tucker, K.~Yang, {\em et~al.}, ``Large scale distributed deep
  networks,'' in {\em Advances in neural information processing systems},
  pp.~1223--1231, 2012.

\bibitem{li2014communication}
M.~Li, D.~G. Andersen, A.~J. Smola, and K.~Yu, ``Communication efficient
  distributed machine learning with the parameter server,'' in {\em Advances in
  Neural Information Processing Systems}, pp.~19--27, 2014.

\bibitem{li2019accelerating}
Y.~Li, I.-J. Liu, Y.~Yuan, D.~Chen, A.~Schwing, and J.~Huang, ``Accelerating
  distributed reinforcement learning with in-switch computing,'' in {\em 2019
  ACM/IEEE 46th Annual International Symposium on Computer Architecture
  (ISCA)}, pp.~279--291, IEEE, 2019.

\bibitem{ho2013more}
Q.~Ho, J.~Cipar, H.~Cui, S.~Lee, J.~K. Kim, P.~B. Gibbons, G.~A. Gibson,
  G.~Ganger, and E.~P. Xing, ``More effective distributed ml via a stale
  synchronous parallel parameter server,'' in {\em Advances in neural
  information processing systems}, pp.~1223--1231, 2013.

\end{thebibliography}

\end{document}